%% file: acl2020.tex
\documentclass[11pt,a4paper]{article}
\usepackage[hyperref]{acl2020}
\usepackage{times}
\usepackage{latexsym}

\usepackage{microtype}

\aclfinalcopy % Uncomment this line for the final submission
\def\aclpaperid{2001} %  Enter the acl Paper ID here

\usepackage{amsmath,graphicx,caption,subcaption,stfloats}
\usepackage{wrapfig}
\newcommand{\Lcal}{\mathcal{L}}

\newcommand{\yccomment}[1]{}
\newcommand{\zhe}[1]{{}}
\newcommand{\JJ}[1]{{}}

\title{Distilling Knowledge Learned in BERT for Text Generation}

\author{Yen-Chun Chen\textsuperscript{\rm 1}, Zhe Gan\textsuperscript{\rm 1}, Yu Cheng\textsuperscript{\rm 1},
Jingzhou Liu\textsuperscript{\rm 2}, Jingjing Liu\textsuperscript{\rm 1} \\
  \textsuperscript{\rm 1}Microsoft Dynamics 365 AI Research \quad \textsuperscript{\rm 2}Carnegie Mellon University\\
\small\texttt{\{yen-chun.chen,zhe.gan,yu.cheng,jinjl\}@microsoft.com; liujingzhou@cs.cmu.edu}  \\}

\date{}

\begin{document}
\maketitle
\begin{abstract}
    Large-scale pre-trained language model such as BERT has achieved great success in language understanding tasks. However, it remains an open question how to utilize BERT for language generation. In this paper, we present a novel approach, Conditional Masked Language Modeling (C-MLM), to enable the finetuning of BERT on target generation tasks. The finetuned BERT (\emph{teacher}) is exploited as extra supervision to improve conventional Seq2Seq models (\emph{student}) for better text generation performance.
    By leveraging BERT's idiosyncratic bidirectional nature, distilling knowledge learned in BERT can encourage auto-regressive Seq2Seq models to plan ahead, imposing global sequence-level supervision for coherent text generation.
    Experiments show that the proposed approach significantly outperforms strong Transformer baselines on multiple language generation tasks such as machine translation and text summarization.
    Our proposed model also achieves new state of the art on IWSLT German-English and English-Vietnamese MT datasets.\footnote{Code is available at \href{https://github.com/ChenRocks/Distill-BERT-Textgen}{https://github.com/ChenRocks/Distill-BERT-Textgen}.}
\end{abstract}

\input{tex/intro}

\input{tex/related}

\input{tex/model}

\input{tex/experiments}
\input{tex/conclusion}

\bibliography{acl2020}
\bibliographystyle{acl_natbib}

\clearpage
\appendix
\input{tex/appendix}

\end{document}

%% file: tex/intro.tex
\section{Introduction}
%
% background
Large-scale pre-trained language model, such as ELMo~\citep{peters2018deep}, GPT~\citep{radford2018improving} and BERT~\citep{devlin2018bert}, has become the \textit{de facto} first encoding step for many natural language processing (NLP) tasks. For example, BERT, pre-trained with deep bidirectional Transformer~\citep{vaswani2017attention} via masked language modeling and next sentence prediction, has revolutionized the state of the art in many language understanding tasks, such as natural language inference~\citep{bowman2015large} and question answering~\citep{rajpurkar2016squad}. 

% motivation
However, beyond common practice of finetuning BERT for language understanding~\citep{wang2018glue}, applying BERT to language generation still remains an open question. Text generation aims to generate natural language sentences conditioned on certain input, with applications ranging from machine translation~\citep{cho2014learning,sutskever2014sequence,bahdanau2014neural}, text summarization~\citep{nallapati2016abstractive,gehring2017convolutional,chen2018fast}, to image captioning~\citep{vinyals2015show,xu2015show,gan2017semantic}. In this work, we study how to use BERT for better text generation, which is still a relatively unexplored territory.

Intuitively, as BERT is learned with a generative objective via Masked Language Modeling (MLM) during the pre-training stage, a natural assumption is that this training objective should have learned essential, bidirectional, contextual knowledge that can help enhance text generation.
Unfortunately, this MLM objective is not auto-regressive, which encumbers its direct application to auto-regressive text generation in practice. 

% method
We tackle this challenge by proposing a novel and generalizable approach to distilling 
knowledge learned in BERT for text generation tasks. We first propose a new Conditional Masked Language Modeling (C-MLM) task, inspired by MLM but requiring additional conditional input, which enables finetuning pre-trained BERT on a target dataset. 
In order to extract knowledge from the finetuned BERT and apply it to a text generation model, we leverage the finetuned BERT as a teacher model that generates sequences of word probability logits for the training samples, and treat the text generation model as a student network, which can effectively learn from the teacher's outputs for imitation. The proposed approach improves text generation by providing a good estimation on word probability distribution for each token in a sentence, consuming both the left and the right context, the exploitation of which encourages conventional text generation models to \emph{plan ahead}.
At inference time, the teacher model (BERT) is not required thus the decoding speed is as fast as the underlying student model.

Text generation models are usually trained via Maximum Likelihood Estimation (MLE), or \emph{teacher forcing}~\citep{bengio2015scheduled}: at each time step, it maximizes the likelihood of the next word conditioned on its previous ground-truth words. This corresponds to optimizing one-step-ahead prediction. As there is no explicit signal towards global planning in the training objective, the generation model may incline to focusing on local structure rather than global coherence.  
With our proposed approach, BERT's \emph{looking into the future} ability can act as an effective regularization method, capturing subtle long-term dependencies that ensure global coherence and in consequence boost model performance on text generation.

An alternative way to leverage BERT for text generation is to initialize the parameters of the encoder or decoder of Seq2Seq with pre-trained BERT, and then finetuning on the target dataset. However, this approach requires the encoder/decoder to be identical to BERT, inevitably making the final text generation model too large. Our approach, on the other hand, is modular and compatible to any text-generation model, and has no restriction on model size or model architecture (e.g., LSTM or Transformer).

% summary of contributions
The main contributions of this work are three-fold:
($i$) We present a novel approach to utilizing BERT for text generation. The proposed method induces sequence-level knowledge into the conventional one-step-ahead and teacher-forcing training paradigm, by introducing an effective regularization term to MLE training loss. 
($ii$) We conduct comprehensive evaluation on multiple text generation tasks, including machine translation and text summarization.
Experiments show that our proposed approach significantly outperforms strong Transformer baselines and is generalizable to different tasks.
($iii$) The proposed model achieves new state of the art on both IWSLT14 German-English and IWSLT15 English-Vietnamese datasets.

%% file: tex/related.tex
\section{Related Work}
\noindent \textbf{Pre-trained Language Models} \, Prior to large-scale pre-trained language model, word embeddings~\citep{mikilov2013distributed,pennington2014glove,bojanowski2017enriching} were widely used for NLP tasks.
Recently, CoVe~\citep{mccann2017learned} introduced (conditional) language models pre-trained on paired machine translation corpus. ELMo~\citep{peters2018deep} learned a contextual language model on a large corpus with bidirectional RNN. GPT~\citep{radford2018improving} used unidirectional Transformer to achieve better contextualized word representation.
By fine-tuning pre-trained language models, ULMFit~\citep{howard-ruder-2018-universal} also achieved promising results on text classification.

In our study, we focus on BERT due to its superior performance on multiple language understanding tasks. However, different from previous work exploiting BERT for language understanding tasks, here we aim to apply BERT to text generation. To the best of our knowledge, this is still a relatively unexplored space. 
The proposed approach is also model-agnostic and can be applied to other pre-trained language models as well.

\vspace{5pt}
\noindent \textbf{BERT for Text Generation}\, There has been some recent attempt on applying BERT to text generation. Specifically,
\citet{lample2019cross} trained cross-lingual MLM and demonstrated promising results for cross-lingual natural language inference~\citep{conneau2018xnli} and unsupervised neural machine translation (NMT)~\citep{lample2017unsupervised}.
\citet{wang2019bert} formulated BERT as a Markov Random Field LM and showed preliminary results on unsupervised text generation with improved diversity.
\citet{zhang2019pretraining} utilized an encoder with BERT and a two-stage decoder for text summarization.
\citet{song2019mass} proposed Masked Seq2Seq (MASS) pre-training, demonstrating promising results on unsupervised NMT, text summarization and conversational response generation.
Concurrent with our work, \citet{ghazvininejad2019constant} proposed a similar conditional MLM for constant-time translation, and \citet{yang2019towards} studied how to fine-tune BERT for NMT.

Our approach is novel in the sense that we do not directly use the parameters of BERT in the Seq2Seq model. Instead, BERT acts as an effective regularization to the MLE training loss, by proactively injecting future information for predicting the present. 

\vspace{5pt}
\noindent \textbf{Right-to-Left Generation}\, Our work also shares a high-level intuition with those approaches that try to regularize left-to-right generative models with a right-to-left counterpart. Specifically, 
\citet{liu-etal-2016-agreement-target} trained a separate reverse NMT and performed joint decoding at inference time to enforce agreement between forward and reverse models.
Twin Networks~\citep{serdyuk2018twin} used a backward RNN jointly trained with a forward RNN decoder by matching their hidden states.
\citet{zhang2018regularizing} further extended the idea to Transformer with joint training, so that the forward and the backward models iteratively improve each other. Our proposed approach stems from a similar intuition. However, we focus on using pre-trained language model such as BERT to regularize an auto-regressive generation model.  

\vspace{5pt}
\noindent \textbf{Knowledge Distillation} \, Our method shares the same loss formulation as Knowledge Distillation (KD) proposed in~\citet{bucilua2006model,hinton2015distilling,kim-rush-2016-sequence}, where a smaller student model is trained on soft labels provided by a larger teacher model.
More recently, \citet{tan2019multilingual} applied KD to multilingual NMT, and \citet{sun2019patient} proposed patient KD for BERT model compression.
Compared with these previous studies, where both the teacher and the student are trained on the same task, our approach is different in the sense that the BERT teacher is not designed to perform the student's generation task.
We focus on using KD to leverage the learned knowledge in BERT for text generation, while previous work mostly focused on model compression.

%% file: tex/model.tex
\input{fig/framework.tex}
\section{Approach}
In this section, we present our proposed approach to distilling the knowledge in BERT for text generation in generic sequence-to-sequence (Seq2Seq) setting. 
We first review Seq2Seq learning in Section~\ref{sec:seq2seq}, and then describe the proposed approach in Section~\ref{sec:cmlm} and \ref{sec:kd}.
%
%\vspace{-5pt}
\subsection{Sequence-to-Sequence Learning} \label{sec:seq2seq}
Seq2Seq learning~\citep{sutskever2014sequence} aims to generate a sequence of discrete output $Y=(y_1,\ldots,y_N)$ of length $N$, conditioned on a sequence of discrete input $X=(x_1,\ldots,x_M)$ of length $M$. A Seq2Seq model learns parameters $\theta$ to estimate the conditional likelihood $P_{\theta}(Y|X)$, typically trained via Maximum Likelihood Estimation (MLE), or equivalently, minimizing the cross-entropy loss: 
%
%\vspace{-10pt}
\begin{align} \label{eqn:clm}
\Lcal_{xe} (\theta) &= -\log P_{\theta}(Y|X) \\
&= -\sum_{t=1}^N \log P_{\theta}(y_t | y_{1:t-1}, X) \,, \nonumber
\end{align}
where each conditional probability can be calculated via an attention-based recurrent neural network (RNN)~\citep{bahdanau2014neural,luong2015effective}, Transformer~\citep{vaswani2017attention}, or any other neural sequence-generation models.

%\vspace{-5pt}
\subsection{Finetune BERT with Conditional MLM}
\label{sec:cmlm}
%\vspace{-5pt}
%

This generic Seq2Seq learning framework is the state of the art on a wide range of text generation tasks. Using modern deep neural networks, the conditional probabilities can be readily modeled as a sequence of classifications over the word vocabulary.
However, during training, in order to generate the $t$-th token $y_t$, the model only sees a partial sentence $y_{1:t-1}$ from the ground-truth training data. Intuitively, it is reasonable to assume that a bidirectional model can be more informative than a left-to-right generation model, since additional context from the right (or future) is also incorporated to predict the current word. Unfortunately, this additional information is not utilized in a standard Seq2Seq model, since it can only be trained in a left-to-right manner, where the future context is masked out to prevent each word from indirectly ``\emph{seeing itself}''. To compensate this single-directional limitation of Seq2Seq setting, we propose a new conditional language model (C-MLM) to enable the finetuning of BERT on target generation task, in hope that the finetuned bidirectional BERT can be utilized for better text generation. 

BERT~\citep{devlin2018bert} is a deep bidirectional Transformer trained via Masked Language Modeling (MLM).\footnote{Besides MLM, \citet{devlin2018bert} also introduced the next sentence prediction task for training BERT. We omit this task since it is unrelated to our work.}
In a similar setting, where the input is a sequence pair ($X, Y$),\footnote{The two sequences are consecutive paragraphs sampled from a very large corpus such as Wikipedia.} $15\%$ of the tokens are randomly masked.
Formally, we denote the masked token sets as $X^m$ and $Y^m$, and the disjoint counterpart (\emph{i.e.}, the unmasked tokens) as $X^u$ and $Y^u$, respectively. 
The trained BERT model aims to estimate the joint probability:
\begin{align} \label{eqn:mlm}
P(x^m_1, \dots, x^m_i, y^m_1, \dots, y^m_j| X^u, Y^u) \,,
\end{align}
where $i$ and $j$ denote the number of masked tokens in $X$ and $Y$, respectively. Each $x^m_\star \in X^m$, and each $y^m_\star \in Y^m$. Eqn. (\ref{eqn:mlm}) can be trained with the standard word-level cross-entropy loss.

We aim to marry MLM pre-training with Seq2Seq learning, to leverage bidirectional language model for text generation.
To this end, we propose a conditional-MLM, a variant of MLM that allows further finetuning of pre-trained BERT on target dataset. For example, for machine translation, $X$ and $Y$ represent the source and the target sentence, respectively. We first concatenate them together and randomly mask $15\%$ of the tokens only in $Y$, then train the network to model the joint probability:
\begin{align}
P(y^m_1, \dots, y^m_j| X, Y^u) \,.
\end{align}
The above C-MLM objective is similar to the conditional language modeling (LM) objective in Eqn. (\ref{eqn:clm}), but conditional LM only permits predicting a word based on its left context. C-MLM is also related to Masked Seq2Seq (MASS) pre-training~\cite{song2019mass}. However, in MASS, the encoder takes a sentence with randomly masked fragment (several consecutive tokens) as input, and the decoder tries to predict this masked fragment, which is different from our model design. The final goal is also different: MASS focuses on Seq2Seq pre-training, while we focus on leveraging BERT for text generation. 
In our experiments, we observe that the C-MLM task can obtain high accuracy and good generalization on word prediction. However, it is not feasible to generate sequential output directly from C-MLM. Instead, we use knowledge distillation to distill the knowledge learned from the finetuned BERT into a Seq2Seq model for direct text generation, which will be explained in the next sub-section.

\subsection{Knowledge Distillation for Generation} \label{sec:kd}
Our inspiration springs from the observation that the probability distribution of the masked word $y_t^m$ is estimated using both  $y_{1:t-1}^u$ and $y_{t+1:N}^u$ from $Y^u$.
In other words, the distribution for a given word $P(y_t^m | X, Y^u)$ contains information from both backward and forward contexts, which is a desirable benefit for providing sequence-level global guidance.
This probability distribution can be considered as soft targets for a text generation model to mimic from, which potentially contains more useful and fine-grained information than the usual hard-assigned, one-hot label, therefore enhancing conventional left-to-right generation models to \emph{look into the future}.\yccomment{} 

In a knowledge distillation setting, the BERT model can be considered as a \emph{teacher}, while the Seq2Seq model acts as a \emph{student}.
Specifically, the Seq2Seq model can be trained with the following objective function:
% KD loss
%\vspace{-10pt}
\begin{align}
\label{eqn:bert_loss}
\mathcal{L}_{bidi}(\theta) = -\sum_{w\in \mathcal{V}}\Big[
    &P_\phi (y_t = w | Y^u, X) \cdot \\
    &\log P_\theta (y_t = w | y_{1:t-1}, X)
\Big] \,, \nonumber
\end{align}
where $P_\phi(y_t)$ is the soft target estimated by the finetuned BERT with learned parameters $\phi$, and $\mathcal{V}$ denotes the output vocabulary. Note that $\phi$ is fixed during the distillation process. An illustration of this learning process is provided in Figure~\ref{fig:framework}, which aims to match the word probability distribution $P_{\theta}(y_t)$ provided by the student with $P_{\phi}(y_t)$ provided by the teacher (\emph{i.e.}, distillation).

To further improve the Seq2Seq student model, hard-assigned labels are also utilized. The final model is trained with the following compound objective:
\begin{align}
\label{eqn:final_loss}
\mathcal{L}(\theta) = \alpha\mathcal{L}_{bidi} (\theta) + (1-\alpha)\mathcal{L}_{xe} (\theta) \,,
\end{align}
where $\alpha$ is a hyper-parameter for tuning the relative importance of the two training targets: soft estimation from finetuned BERT, and ground-truth hard label.
Note that our proposed approach only has a minimal requirement on the architecture of the incorporated Seq2Seq model.
As long as the model is trained to estimate word-level probability as in~Eqn. (\ref{eqn:clm}), it can be trained jointly with the proposed objective function~Eqn. (\ref{eqn:final_loss}).

At a higher level, the additional loss term $\mathcal{L}_{bidi}$ can be interpreted as a sequence-level objective function.
Our auto-regressive (or causal) model $\theta$ tries to predict the probability distribution that matches the estimation the bidirectional teacher model predicts, hence encouraging the planning of future (right context) for generation.

%% file: fig/framework.tex
\begin{figure*}
 \centering
 \includegraphics[width=0.8\textwidth]{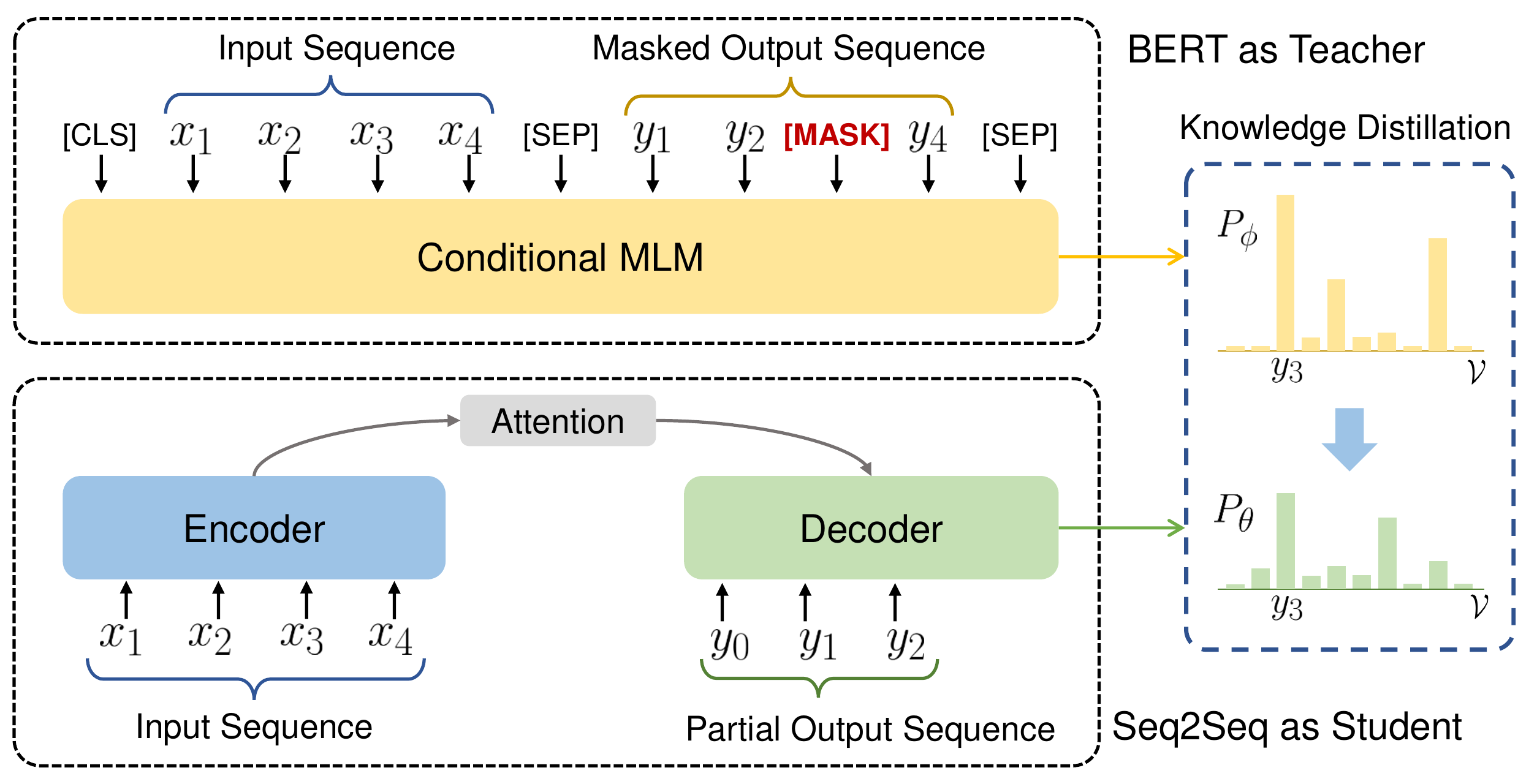}
 \vspace{-10pt}
 \caption{Illustration of distilling knowledge from BERT for text generation. See Section~\ref{sec:cmlm} and \ref{sec:kd} for details.}
 \label{fig:framework}
 %\vspace{-15pt}
\end{figure*}

%% file: tex/experiments.tex
\section{Experiments}
In this section, we describe our experiments on two well-studied text generation tasks: machine translation, and abstractive text summarization.
\subsection{Datasets}
\label{subsec:dataset}
\textbf{Machine Translation}\,
% small-scale
We consider two relatively small-scale datasets, IWSLT15 English-Vietnamese (En-Vi, 113k training samples) and IWSLT14 German-English (De-En, 160k training samples), and one medium-scale dataset, WMT14 English-German (En-De, 4.5M training samples).
For IWSLT15 En-Vi, we use the pre-processed dataset provided by~\citet{Luong-Manning:iwslt15}.
We use tst2012 as dev set and test on tst2013.
For IWSLT14 De-En, we follow the pre-processing steps and the same train/dev/test split as in~\citet{wu2018pay}.
For WMT14 En-De, we follow the pre-processing steps in~\citet{vaswani2017attention} for fair comparison. We use newstest2013 as the dev set and newstest2014 as the test set.
We report BLEU scores~\citep{papineni-etal-2002-bleu} for evaluation of MT performance following the Moses script.\footnote{For fair comparison to previous work, we report tokenized BLEU scores using
\href{https://github.com/moses-smt/mosesdecoder/blob/master/scripts/generic/multi-bleu.perl}{https://github.com/moses-smt/mosesdecoder/blob/master/scripts/generic/multi-bleu.perl},
and for WMT14 En-De, we further split the compound words after tokenization.}

\vspace{5pt}
\noindent \textbf{Abstractive Summarization}\,
For summarization, we conduct experiments on the Gigaword summarization dataset~\citep{rush-etal-2015-neural}.
Note that the original train/valid/test split of Gigaword is 3.8M/190k/2k.
In our experiments, we observed severe distribution mismatch between the validation and test data. See Table~\ref{tab:giga-dev},~\ref{tab:giga-test}, and Sec.~\ref{sec:summ-result} for detailed discussion.
Therefore, we further sampled 5k/5k dev/test-dev splits from the validation set and tuned hyper-parameters on the dev set only.
We report ROUGE scores~\citep{lin-2004-rouge} on test-dev for the evaluation of our proposed approach, and include results on the standard test split for the comparison with prior work.

\subsection{Implementation Details}
Our implementation is based on the PyTorch~\citep{paszke2017automatic} version of OpenNMT~\citep{klein-etal-2018-opennmt} seq2seq toolkit.
We use the `base' model of 6-layer Transformer with 512-hidden 8-head attention blocks and 2048-hidden feed-forward layer for all experiments, with label smoothing regularization (LSR)~\citep{szegedy2016rethinking} of 0.1.\footnote{Our method can also be viewed as a `learned LSR'. The results reported of our proposed method are trained together with regular LSR, showing the effectiveness of our teacher.}
We batch examples with similar sequence length, and count batch size by the number of tokens.
For MT we use the pre-trained \textit{BERT-base-multilingual-cased} model, and for summarization we use \textit{BERT-base-uncased} as the starting point of BERT finetuning.\footnote{BERT pre-trained models are available at \href{https://github.com/google-research/bert}{https://github.com/google-research/bert}. Our finetuning implementation is modified from code available at \href{https://github.com/huggingface/pytorch-pretrained-BERT}{https://github.com/huggingface/pytorch-pretrained-BERT}.}
We use the corresponding pre-trained byte-pair-encoding~\citep{sennrich-etal-2016-neural} shipped together with the BERT model for tokenization.

For all training methods of all Transformer models, the learning rate schedule is set to
$lr = \eta \cdot d^{-0.5}_{model}\cdot \min(step^{-0.5}, step\cdot warmup\_steps^{-1.5}),$
where $d_{model} = 512$ is the attention representation size~\citep{vaswani2017attention}.
For all BERT finetuning, we follow \citet{devlin2018bert} and use a triangular learning rate schedule with maximum learning rate $\eta$.
The parameters are updated with the Adam optimizer~\citep{kingma2015adam}.
In the distillation stage, we pre-compute BERT's prediction logits of the training data\footnote{The masking strategy is described in the supplementary.} and use top-$K$ distillation~\citep{tan2019multilingual} to reduce computation overhead and memory footprint,
where $K$ is set to 8 across all the experiments.\footnote{We also tune the temperature $T$ for the $softmax$ applied at the teacher's logits. Different from the original KD, we do not apply the same $T$ on the student. In preliminary experiment we found high $T$ of Seq2Seq results in much worse performance. We hypothesize the low-entropy nature of conditioned text generation is not suitable for temperature scaling.}

For the detailed values of the hyper-parameters for each experiment, please refer to the supplementary material.
We found it necessary to train longer with $\mathcal{L}_{bidi}$, since it is still improving after the step at which the baseline Transformer starts to plateau.
At inference time, we use beam search with beam size 4 and length penalty~\citep{wu2016gnmt} of 0.6 across all the models.
All the hyper-parameters are tuned on the development set.
Note that our Transformer baselines achieve higher scores than the reference implementation on each dataset (in most cases comparable to the state-of-the-art).

\input{table/deen}

\input{table/envi}

\input{table/ende}

%\vspace{-5pt}
\subsection{Results on Machine Translation}
We first validate our proposed text generation approach on machine translation task. Experimental results are summarized in Table~\ref{tab:deen},~\ref{tab:envi} and~\ref{tab:ende}, which show that our model significantly improves over the strong Transformer baseline across all three datasets.
Note that our baseline is the `base' model of Transformer, which has 44M trainable parameters, and the reference implementation by~\citet{wu2018pay} of the `big' model with 176M parameters.\footnote{Parameter counts exclude word embedding and final linear projection, which mostly depends on the vocabulary size. BERT-base has 86M trainable parameters.}

For IWSLT German-English translation, our method improves over the Transformer baseline by 1.54 BLEU points, and achieves new state of the art.
Our approach outperforms previously-reported results such as ConvS2S+MRT, a convolutional-based model~\citep{gehring2017convolutional} with minimum risk training~\citep{edunov-etal-2018-classical}, and Lightweight and Dynamic Convolution~\citep{wu2018pay}.
Note that ~\citet{wu2018pay} also tuned checkpoint averaging, which creates a soft ensemble effect. And their model has roughly the same amount of parameters as Transformer (big).

For IWSLT English-Vietnamese translation, since most prior work experimented with RNN models, we also report RNN-based results here.
This also suggests that our method is model-agnostic.
Our best model outperforms Seq2Seq-OT~\citep{chen2019improving} that utilizes optimal transport for sequence-level training, as well as the ELMo and CVT results reported in~\citet{Clark2018CVT}.\footnote{The CVT results used a much larger RNN and CNN-based character embedding, as well as a customized structure. Therefore, we did not try to use RNN to match their results. }
For WMT14 English-German translation, our method still improves over the well-tuned Transformer baseline.
We also report the scores of Transformer (big) and state-of-the-art Dynamic Convolution model~\citep{wu2018pay} for reference.

\subsection{Results on Abstractive Summarization}
\label{sec:summ-result}

Table~\ref{tab:giga-dev} and Table~\ref{tab:giga-test} show the results of our approach on abstractive summarization task, where R-1, R-2, and R-L denote $F_1$ scores of ROUGE-1, ROUGE-2, and ROUGE-L, respectively.
Our method shows improvement on all the metrics, as shown in Table~\ref{tab:giga-dev}.
We observe a large gap between dev and test scores, which suggests that the data in the test set is very different from that in the validation set, as mentioned in Section~\ref{subsec:dataset}.
Given the fact that the official test split contains only 1,951 noisy examples,\footnote{
When we manually inspected the test set data, we found many corrupted examples such as extremely short input articles, meaningless summary, and dominating unknown words.}
we believe that our results on the dev/test-dev sets further strengthens our claim.

\input{table/giga}

\input{table/giga_test}

On the test split, our best model is comparable to state-of-the-art models that use much more complex architectures specifically designed for summarization.
CGU~\citep{lin-etal-2018-global} augmented convolutional gating units.
FTSum$_g$~\citep{Cao2018Faithful} leveraged extra information extraction and dependency parsing features.
E2T$_{cnn}$~\citep{amplayo-etal-2018-entity} utilized entities provided by an external entity linking system.
Re$^3$Sum~\citep{cao-etal-2018-retrieve} carefully designed a retrieve-and-rerank pipeline with human-written soft templates.
Despite that our model has no summarization-specific model design, we still achieve comparable performance to these models on all the metrics.

\subsection{Ablation Study}
There are several possible factors that could contribute to the performance gain: additional parameters of BERT, extra data (pretraining corpus) of BERT, and the bidirectional nature.
To better understand the key contributions of our method, we conduct an ablation study described in the following.
We finetune 2 extra teachers: BERT$_{sm}$ and BERT$_{l2r}$.
For BERT$_{sm}$, we use a smaller BERT (6 layers) for C-MLM finetuning, which has approximately the same number of parameters as Transformer-base.\footnote{We still use the pretrained weights of BERT, otherwise the C-MLM does not converge very well.}
For BERT$_{l2r}$, we use the full BERT model but finetune it using left-to-right LM as in the conventional Seq2Seq model.
Next, we apply the proposed KD method to train the Transformer on En-Vi and De-En MT tasks.
Results are shown in Table~\ref{tab:ablation}.
BERT$_{sm}$ still works well though the full BERT provides further improvement.
On the other hand, BERT$_{l2r}$ slightly hurts the performance.
We hypothesize that it generates noisy learning targets for the student, hence the performance drop.
Empirically, we show that the bidirectional knowledge could be more important than the extra parameters, while the pre-trained weights remain useful for more stable C-MLM training.

\input{table/ablation.tex}
\input{fig/len-deen_envi.tex}
\input{fig/len-ende.tex}

\input{table/examples.tex}

\subsection{Generation for Different Lengths}
We next analyze the effect of our proposed approach on different output lengths.
We plot the BLEU scores on MT w.r.t. different output generation lengths $N$ on the development set.\footnote{For Gigaword summarization, almost all summaries are short sentences (less than 0.5\% of the summaries contain more than 16 words), so we omit the analysis.}
Results are provided in Figure~\ref{fig:len-deen_envi} and Figure~\ref{fig:len-ende}. 
For IWSLT German-English dataset (Figure~\ref{fig:len-deen_envi}: Left), we can see a shared trend that the proposed $\mathcal{L}_{bidi}$ objective gains higher BLEU points on longer translation pairs.
For WMT English-German (Figure~\ref{fig:len-ende}), we can see that although the proposed method performs much worse when the output sentences are very short, it achieves relatively consistent improvement on longer cases, hence resulting in overall BLEU improvement.
For IWSLT English-Vietnamese (Figure~\ref{fig:len-deen_envi}: Right), we see a similar trend when the length $N > 24$.

\subsection{Qualitative Examples}
\label{ssec:examples}
In Table~\ref{tab:examples}, we show some translation examples  on IWSLT German-English dataset.
In the first example, the baseline Transformer cannot recover from `\emph{with}' and `\emph{of}', which renders the full sentence not making much sense.
``I started reading \emph{with}...'' would make sense from the left context; however, if the model also considers the right context ``the age of two'', the word `\emph{with}' would be assigned with lower probability by the soft labels provided by the BERT teacher.
Even though at test-time the model cannot `look ahead', the soft-targets at training-time prevents the over-confidence of the model on one-hot label; hence the better generalization at the test-time.
Similarly, other examples show that our model can generate text more coherently w.r.t. the context on the right (\underline{underlined} in Table~\ref{tab:examples}), thus making more accurate and natural translation.

%% file: table/deen.tex
\begin{table}[t!]
%\begin{small}
\centering
\begin{tabular}{ | p{0.29\textwidth} | c  c|}
  \hline
  De-En Models & dev & test \\
  \hline
  \multicolumn{3}{|c|} {Our Implementations} \\
  \hline
  Transformer (base) & 35.27 & 34.09 \\
     + BERT teacher & \textbf{36.93} & \textbf{35.63} \\
     %+ BERT rerank & TODO & TODO \\
     %+ avg + lp & TODO$\dagger$ & TODO$\dagger$ \\
  \hline
  \multicolumn{3}{|c|} {Other Reported Results} \\
  \hline
  ConvS2S + MRT$^\ddagger$ & 33.91 & 32.85 \\
  Transformer (big)$^\diamond$ & - & 34.4$^\dagger$ \\
  Lightweight Conv$^\diamond$ & - & 34.8$^\dagger$ \\
  Dyn. Convolution$^\diamond$ & - & 35.2$^\dagger$ \\
  \hline
\end{tabular}
\vspace{-5pt}
\caption{BLEU scores for IWSLT14 German-English translation.
($\dagger$) tuned with checkpoint averaging.
($\ddagger$) from~\citet{edunov-etal-2018-classical}. ($\diamond$) from~\citet{wu2018pay}.
}
\label{tab:deen}
%\vspace{-5pt}
%\end{small}
\end{table}

%% file: table/envi.tex
\begin{table}[t]
%\begin{small}
\centering
\begin{tabular}{ | p{0.25\textwidth} | c  c|}
  \hline
  En-Vi Models & tst2012 & tst2013 \\
  \hline
  \multicolumn{3}{|c|} {Our Implementations} \\
  \hline
  RNN & 23.37 & 26.80 \\
     + BERT teacher & 25.14 & 27.59 \\
  Transformer (base) & 27.03 & 30.76 \\
     + BERT teacher & \textbf{27.85} & \textbf{31.51} \\
     %+ BERT rerank & TODO & TODO \\
  \hline
  \multicolumn{3}{|c|} {Other Reported Results} \\
  \hline
  RNN$^\dagger$ & - & 26.1 \\
  Seq2Seq-OT$^\star$ & 24.5 & 26.9 \\
  ELMo$^\diamond$ & - & 29.3 \\
  CVT$^\diamond$ & - & 29.6 \\
  \hline
\end{tabular}
\vspace{-5pt}
\caption{BLEU scores for IWSLT15 English-Vietnamese translation. 
($\dagger$) from~\citet{luong17}.
($\star$) from~\citet{chen2019improving}.
($\diamond$) from~\citet{Clark2018CVT}.
}
\vspace{-4pt}
\label{tab:envi}
%\end{small}
\end{table}

%% file: table/ende.tex
\begin{table}[t]
\centering
\begin{tabular}{ | p{0.23\textwidth} | c  c|}
  \hline
  En-De Models & NT2013 & NT2014 \\
  \hline
  \multicolumn{3}{|c|} {Our Implementations} \\
  \hline
  Transformer (base) & 25.95 & 26.94 \\
     + BERT teacher & \textbf{26.22} & \textbf{27.53} \\
  %Transformer (big) & 26.17 & 27.55 \\
  %   + BERT teacher & \textbf{26.45} & \textbf{28.05} \\
  \hline
  \multicolumn{3}{|c|} {Other Reported Results} \\
  \hline
  Transformer (base)$^\diamond$ & 25.8 & 27.3$^\dagger$ \\
  Transformer (big)$^\star$$^\ddagger$ & 26.5 & 29.3$^\dagger$ \\
  Dyn. Convolution$^\bullet$$^\ddagger$ & \textbf{26.9}\small{$\pm$0.2} & \textbf{29.7}$^\dagger$ \\
  \hline
\end{tabular}
\caption{BLEU scores for WMT14 English-German translation.
($\dagger$) tuned with checkpoint averaging. ($\ddagger$) trained on WMT16, a slightly different version of training data.
($\diamond$) from~\citet{vaswani2017attention}.
($\star$) from~\citet{ott2018scaling}.
($\bullet$) from~\citet{wu2018pay}.
}
\label{tab:ende}
\vspace{-15pt}
\end{table}

%% file: table/giga.tex
\begin{table}[t]
\centering
\begin{tabular}{ | p{0.21\textwidth} | c  c  c|}
  \hline
  GW Models & R-1 & R-2 & R-L \\
  \hline
  \multicolumn{4}{|c|} {Dev} \\
  \hline
  Transformer (base) & 46.64 & 24.37 & 43.17 \\
     + BERT teacher & \textbf{47.35} & \textbf{25.11} & \textbf{44.04} \\
     %+ BERT rerank & TODO & TODO & TODO \\
  \hline
  \multicolumn{4}{|c|} {Test-Dev} \\
  \hline
  Transformer (base) & 46.84 & 24.80 & 43.58 \\
     + BERT teacher & \textbf{47.90} & \textbf{25.75} & \textbf{44.53} \\
     %+ BERT rerank & TODO & TODO & TODO \\
  \hline
\end{tabular}
\vspace{-5pt}
\caption{ROUGE F$_1$ scores for Gigaword abstractive summarization on our internal test-dev split.}
%\vspace{-8pt}
\label{tab:giga-dev}
\end{table}

%% file: table/giga_test.tex
\begin{table}[t]
\centering
\begin{tabular}{ | p{0.21\textwidth} | c  c  c|}
  \hline
  GW Models & R-1 & R-2 & R-L \\
  \hline
  Seq2Seq$^\dagger$ & 36.40 & 17.77 & 33.71 \\
  CGU$^\ddagger$ & 36.3 & 18.0 & 33.8 \\
  FTSum$_g$$^\star$ & 37.27	& 17.65 & 34.24 \\
  E2T$_{cnn}$$^\diamond$ & 37.04 & 16.66 & \textbf{34.93} \\
  Re$^3$Sum$^\bullet$ & 37.04 & \textbf{19.03} & 34.46 \\
  \hline
  %Transformer (ours) & 37.00 & 18.46 & 34.49 \\
  Trm + BERT teacher & \textbf{37.57} & 18.59 & 34.82 \\
  \hline
\end{tabular}
\vspace{-5pt}
\caption{ROUGE F$_1$ scores for Gigaword abstractive summarization on the official test set (Trm: Transformer).
($\dagger$) from~\citet{nallapati2016abstractive}.
($\ddagger$) from~\citet{lin-etal-2018-global}.
($\star$) from~\citet{Cao2018Faithful}.
($\diamond$) from~\citet{amplayo-etal-2018-entity}.
($\bullet$) from~\citet{cao-etal-2018-retrieve}.
}
%\vspace{-15pt}
\label{tab:giga-test}
\end{table}

%% file: table/ablation.tex
\begin{table}[t]
\centering
\begin{tabular}{ | p{0.23\textwidth} | c  c|}
  \hline
  Methods & De-En & En-Vi \\
  & (dev) & (tst2012) \\
  \hline
  Transformer (base) & 35.27 & 27.03 \\
  Trm + BERT$_{l2r}$ & 35.20 & 26.99 \\
  Trm + BERT$_{sm}$ & 36.32 & 27.68 \\
  \hline
  Trm + BERT & \textbf{36.93} & \textbf{27.85} \\
  \hline
\end{tabular}
\vspace{-5pt}
\caption{Ablation study. (Trm: Transformer)
}
%\vspace{-10pt}
\label{tab:ablation}
\end{table}

%% file: fig/len-deen_envi.tex
\begin{figure*}[t]
    \centering
    \includegraphics[width=0.49\textwidth]{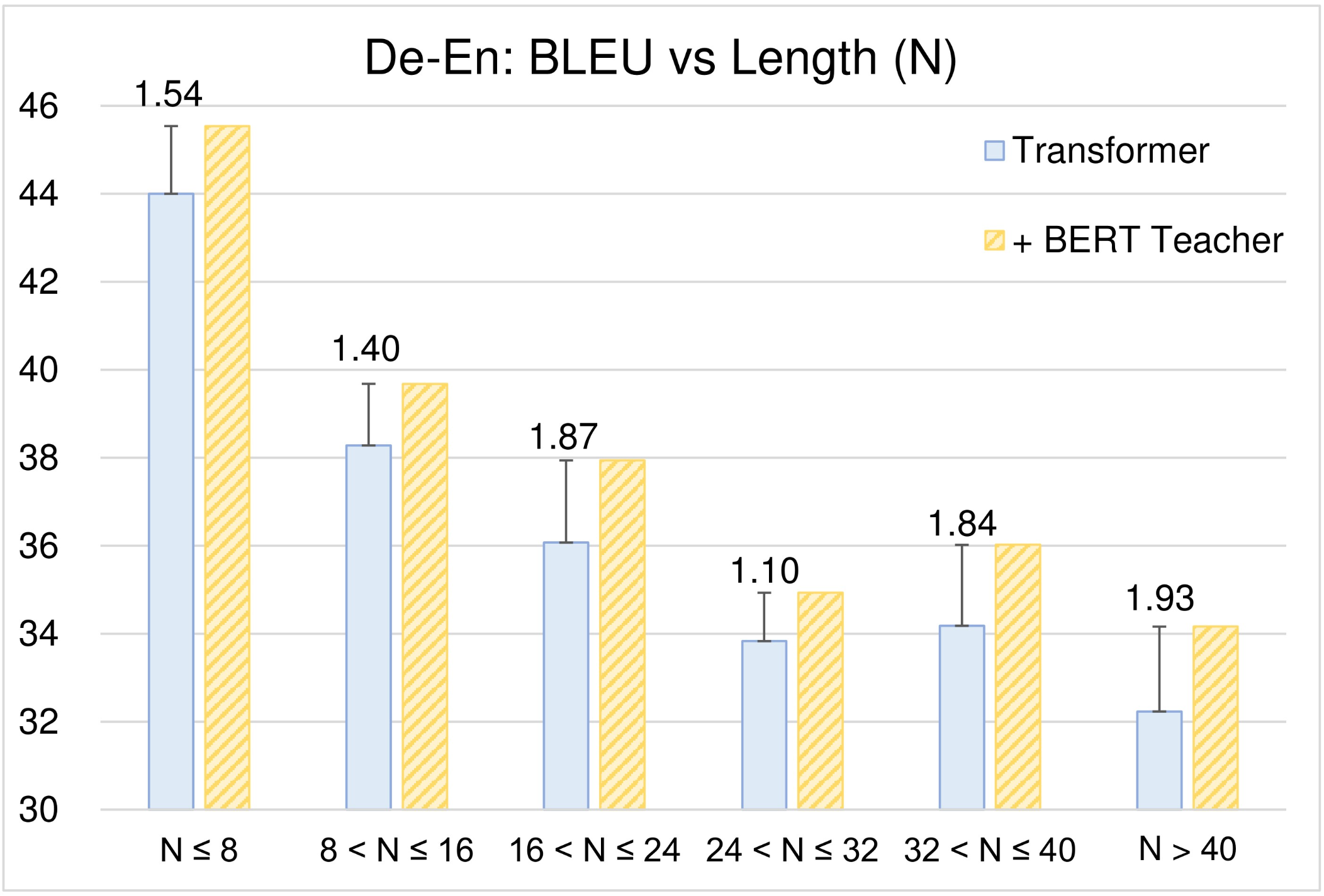}
    \includegraphics[width=0.49\textwidth]{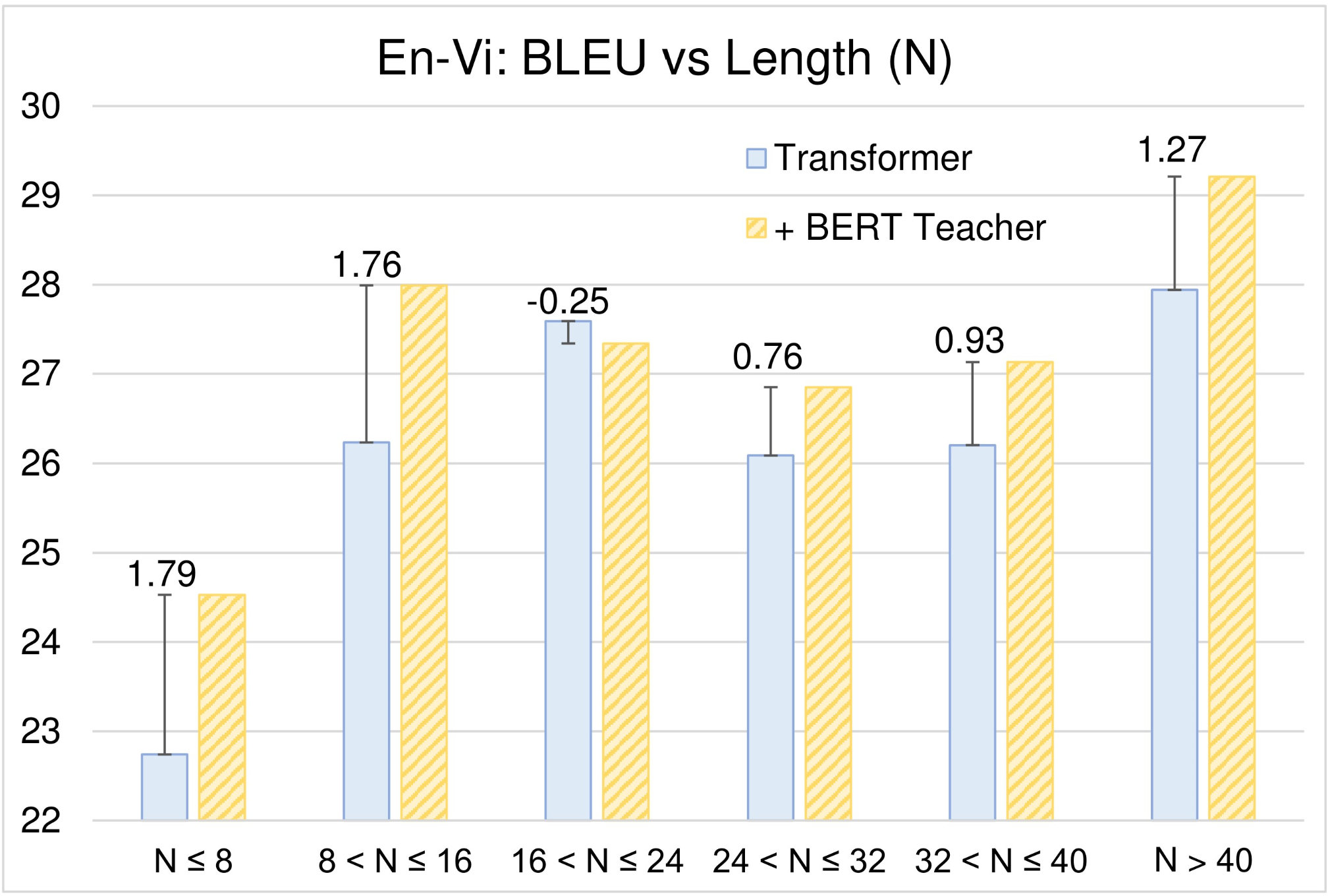}
    \vspace{-5pt}
    \caption{BLEU scores on IWSLT German-English and English-Vietnamese for different output lengths.}
    \label{fig:len-deen_envi}
\end{figure*}

%% file: fig/len-ende.tex
\begin{figure}[t]
    \centering
    \includegraphics[width=0.48\textwidth]{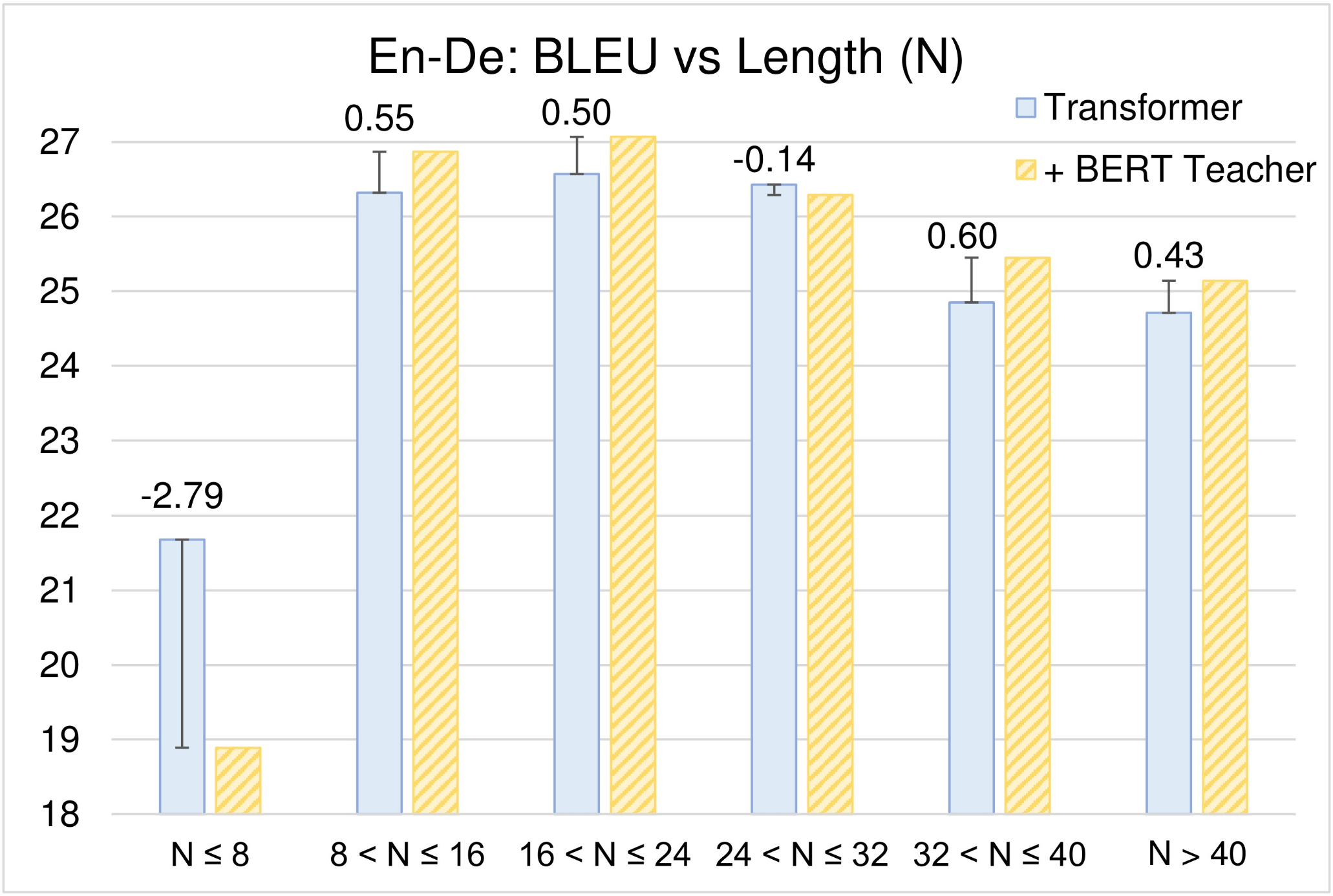}
    \vspace{-5pt}
    \caption{BLEU scores on WMT English-German for different output lengths.}
    \label{fig:len-ende}
    \vspace{-5pt}
\end{figure}

%% file: table/examples.tex
\begin{table*}[t!]
\centering
\begin{small}
\begin{tabular}{  l  p{0.85\textwidth} }
%\hline
%Reference &  
%the political climate in the u.s. at the time was tense , and there were debates going on about immigration . \\
%Transformer &  
%the political climate in the u.s. was back then , and there was constant disasters . (29.5)\\
%Ours &  
%the political climate in the united states at the time was tense , and there were ongoing shifting debates . (57.3) \\
\hline
Reference & my mother says that i started reading at the age of two , although i think four is probably close to the truth . \\
Transformer & my mother says that i started reading \underline{\textcolor{red}{with} two years} , but i think that four \underline{\textcolor{red}{of} them} probably correspond to the truth . (39.6) \\
Ours & my mother says that i started reading \underline{\textcolor{blue}{at} the age of two} , but i think four \underline{\textcolor{blue}{is} more likely to be the truth }. (65.2) \\
%\vspace{-10pt} \\
\hline

Reference & we already have the data showing that it reduces the duration of your flu by a few hours . \\
Transformer & we 've already got the data showing that it 's going to \underline{\textcolor{red}{crash} the duration} of your flu by a few hours . (56.6)\\
Ours & we already have the data showing that it \underline{\textcolor{blue}{reduces} the duration} of your flu by a few hours . (100.0)\\
%\vspace{-10pt} \\
\hline

Reference & we now know that at gombe alone , there are nine different ways in which chimpanzees use different objects for different purposes . \\
Transformer & we know today that alone in gombe , there are nine different ways that chimpanzees use different objects \underline{\textcolor{red}{in} different ways} . (35.8) \\
Ours &  we now know that in gombe alone , there are nine different ways that chimpanzees use different objects \underline{\textcolor{blue}{for} different purposes} . (71.5) \\
%\vspace{-10pt} \\
\hline

\end{tabular}
\end{small}
\vspace{-5pt}
\caption{Qualitative examples from IWSLT German-English translation. 
Numbers inside the parenthesis are sentence-level BLEU scores.
\textcolor{red}{Red} word is where the baseline Transformer makes a mistake without considering the possible \underline{future phrase} and fails to recover.
On the other hand, our model makes the right decision at the \textcolor{blue}{blue} word, hence generates more coherent sentence.
Please refer to Section~\ref{ssec:examples} for detailed explanation.}
\label{tab:examples}
%\vspace{-10pt}
\end{table*}

%% file: tex/conclusion.tex
%\vspace{-5pt}
\section{Conclusion}
%\vspace{-5pt}
In this work, we propose a novel and generic approach to utilizing pre-trained language models to improve text generation \emph{without explicit parameter sharing, feature extraction, or augmenting with auxiliary tasks.}
Our proposed Conditional MLM mechanism leverages unsupervised language models pre-trained on large corpus, and then adapts to supervised sequence-to-sequence tasks.
Our distillation approach \emph{indirectly} influences the text generation model by providing soft-label distributions only, hence is \emph{model-agnostic}.
Experiments show that our model improves over strong Transformer baselines on multiple text generation tasks such as machine translation and abstractive summarization, and achieves new state-of-the-art on some of the translation tasks.
For future work, we will explore the extension of Conditional MLM to multimodal input such as image captioning.

%% file: tex/appendix.tex
\section{Implementaion Details and Hyper-parameter Values}
We run all experiments on single GPU of NVIDIA Titan RTX or V100 except for WMT En-De we use 4 V100s for training.
Note that for large batch sizes that do not fit in GPU memory, we use the gradient accumulation tricks as in \citet{ott2018scaling}.
Batch sizes are counted in number of tokens.
Note that all the hyper-parameters are tuned on the development set only.

To compute the logits (soft labels) from teacher, we repeat a training pair for 7 times and create a circular mask as illustrated in Figure~\ref{fig:mask}. This mask approximates the $15\%$ masking rate of the BERT training. From the masked positions we can obtain soft probabilities predicted by the BERT teacher for each output tokens $y$. These logits are pre-computed once for the training set so that we do not have to repeatedly sample random masks and run forward pass of BERT while training.

\paragraph{IWSLT De-En}
For C-MLM fine-tuning, we train for 100k steps with 5k $warmup\_steps$, $\eta = 5\cdot10^{-5}$, and batch size of 16k tokens.
For baseline model, we train for 50k steps with 4k $warmup\_steps$ and batch size of 6k tokens.
The learning rate $\eta$ is set to 1.
For the proposed model, we train for 100k steps with 8k $warmup\_steps$ and batch size of 6k tokens.
The learning rate $\eta$ is set to 2, $\alpha = 0.5$, and $T = 10$.
Seq2Seq model uses dropout~\citep{srivastava2014dropout} of 0.3 in both cases.
\paragraph{IWSLT En-Vi}
For C-MLM fine-tuning and baseline Transformer, the hyper-parameters are identical to that of IWSLT De-En.
For the proposed model, we train for 100k steps with 8k $warmup\_steps$ and batch size of 6k tokens.
The learning rate $\eta$ is set to 2, $\alpha = 0.1$, and $T = 5$.
Dropout is still 0.1.
\paragraph{WMT En-De}
For C-MLM fine-tuning, we train for 100k steps with 5k $warmup\_steps$, $\eta = 5\cdot10^{-5}$, and batch size of 512k tokens.
For baseline model, we train for 30k steps with 4k $warmup\_steps$ and batch size of 384k tokens.
The learning rate $\eta$ is set to 4.
Since this is our largest dataset and training is slow, for the proposed model we use the baseline Transformer to initialize the Seq2Seq student.
For the proposed model, we continue training for 50k steps with 4k $warmup\_steps$ and batch size of 64k tokens.
The learning rate $\eta$ is set to 2, $\alpha = 0.1$, and $T = 5$.
Seq2Seq model uses dropout of 0.1 in both cases.
\paragraph{Gigaword}
For C-MLM fine-tuning, we train for 100k steps with 5k $warmup\_steps$, $\eta = 5\cdot10^{-5}$, and batch size of 64k tokens.
For baseline model, we train for 50k steps with 4k $warmup\_steps$ and batch size of 40k tokens.
The learning rate $\eta$ is set to 1.
For the proposed model, we train for 70k steps with 4k $warmup\_steps$ and batch size of 36k tokens.
The learning rate $\eta$ is set to 2, $\alpha = 0.1$, and $T = 10$.
Seq2Seq model uses dropout of 0.1 in both cases.

\input{fig/mask.tex}
\input{table/examples-2.tex}

\input{table/examples-summ.tex}

\section{Additional Generation Examples}
We show Gigaword summarization examples in Table~\ref{tab:examples-summ} and extra En-DE generation examples in Table~\ref{tab:examples-2}.
Qualitatively, our Transformer + BERT Teacher outperforms baseline Transformer and generate more coherent sentences.

%% file: fig/mask.tex
\begin{figure}[t]
 \centering
 \includegraphics[width=0.48\textwidth]{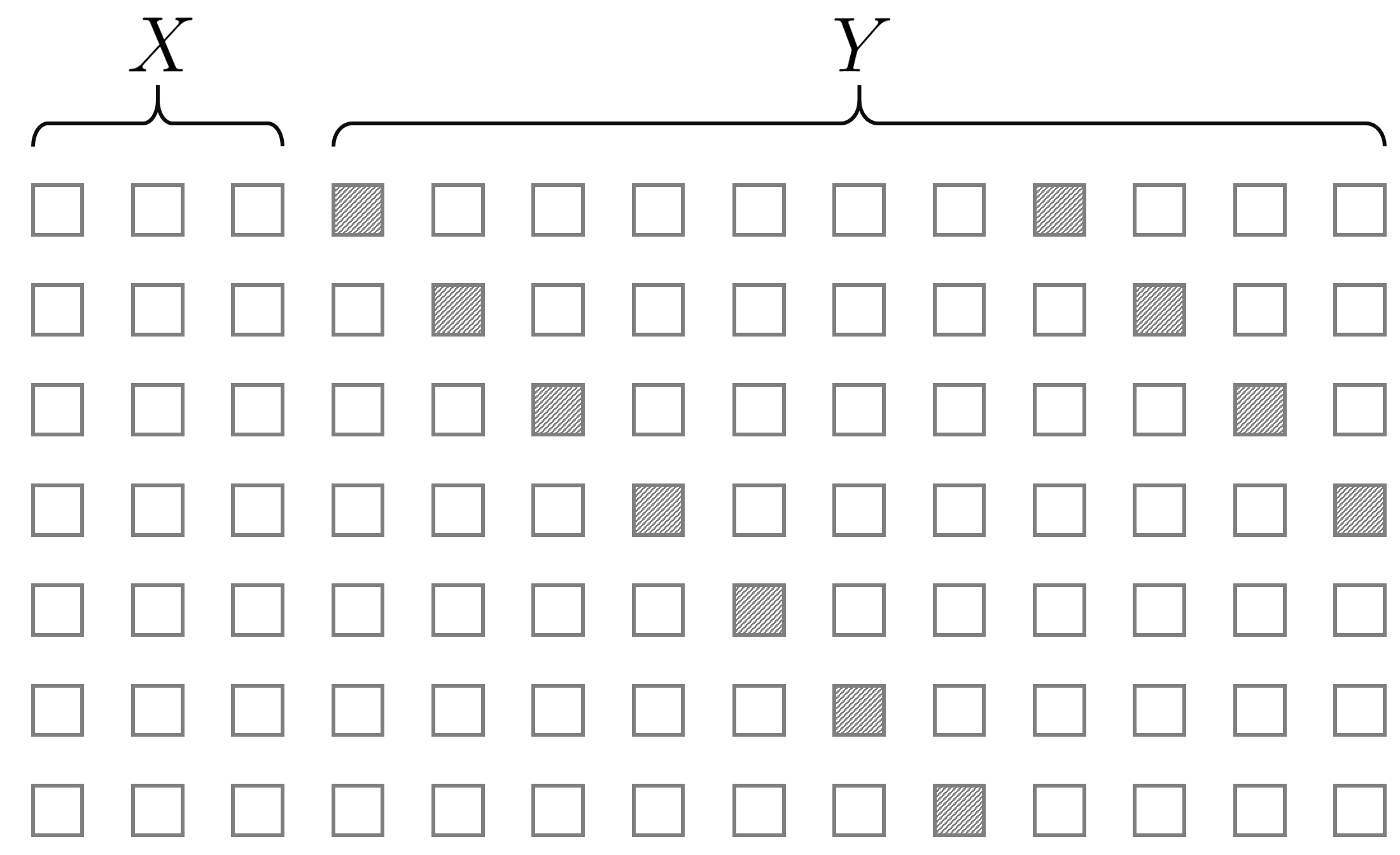}
 \caption{Illustration of the masking strategy for computing the teacher soft labels. Gray slashed boxes denote the \texttt{[MASK]} positions.}
 \label{fig:mask}
\end{figure}

%% file: table/examples-2.tex
\begin{table*}[tp]
\centering
\begin{small}
\begin{tabular}{  l  p{0.85\textwidth} }
\hline
Reference &  
the political climate in the u.s. at the time was tense , and there were debates going on about immigration . \\
Transformer &  
the political climate in the u.s. was \underline{\textcolor{red}{back} then , and there was constant disasters} . (29.5)\\
Ours &  
the political climate in the united states at the time was \underline{\textcolor{blue}{tense} , and there were ongoing shifting debates} . (57.3) \\
\vspace{-10pt} \\
\hline
Reference & it would be immoral to leave these young people with a climate system spiraling out of control . \\
Transformer & it would be immoral to \underline{\textcolor{red}{let} these young people leave a climate system} that was out of control . (44.6)\\
Ours & it would be immoral to \underline{\textcolor{blue}{leave} these young people with a climate system} out of control . (84.3)\\
\vspace{-10pt} \\
\hline
%Reference & and this man with a planetary-sized brain is now serving a 13-year sentence in california . \\
%Transformer & and this man , with a planet-sized brain , is now a 13-year-\textcolor{red}{old} sentence in california . (26.3) \\
%Ours & and this man , with a planet-sized brain , is now a 13-year sentence in california . (42.6) \\
%\vspace{-10pt} \\
%\hline
Reference & the tahltan have called for the creation of a tribal heritage reserve which will set aside the largest protected area in british columbia . \\
Transformer & tahltan demands the \textcolor{red}{institution} of a tribe in british columbia that should make the largest \underline{protection area} in british columbia . (19.9) \\
Ours & the tahltan demands to \textcolor{blue}{build a tribe reserve} that should be the largest \underline{protected area} in british columbia . (32.2) \\
\vspace{-10pt} \\
\hline

\end{tabular}
\end{small}
\vspace{-5pt}
\caption{Qualitative examples from IWSLT German-English translation. 
Numbers inside the parenthesis are sentence-level BLEU scores.
\textcolor{red}{Red} word is where the baseline Transformer makes a mistake without considering the possible \underline{future phrase} and fails to recover.
On the other hand, our model makes the right decision at the \textcolor{blue}{blue} word, hence generates more coherent sentence.
Please refer to Section~4.6 in the main paper for detailed explanation.
%Please refer to Section~\ref{ssec:examples} for detailed explanation.
}
\label{tab:examples-2}
\vspace{-10pt}
\end{table*}

%% file: table/examples-summ.tex
\begin{table}[h]
\renewcommand*{\arraystretch}{1.3}
\centering
\begin{small}
\begin{tabular}{  l  p{0.33\textwidth} }
\hline
Reference  & china offers tax exemptions for laid-off workers \\
Transformer &  china encourages laid-off workers to seek employment \\
Ours & china offers tax exemptions to laid-off workers \\
\vspace{-10pt} \\
\hline
Reference & swiss police arrest britons who allegedly ran rental car racket \\
Transformer &  three britons arrested in swiss luxury hotel \\
Ours & swiss police arrest three britons in rental car racket case \\
\vspace{-10pt} \\
\hline
Reference & south korea stocks extend declines as kia concerns intensify \\
Transformer & south korean stocks fall for \#th time in \# days ; kia leads \\
Ours & south korean stocks fall as kia troubles intensify \\
\vspace{-10pt} \\
\hline

\end{tabular}
\end{small}
\caption{Qualitative examples from the Gigaword summarization dataset. Baseline model suffers from early mistakes. Our model generates more coherent summaries.}
\label{tab:examples-summ}
\end{table}